\documentclass[letterpaper]{article} 
\usepackage{aaai2026}
\usepackage{times}  
\usepackage{helvet}  
\usepackage{courier}  
\usepackage[hyphens]{url}  
\usepackage{graphicx} 
\usepackage{natbib}  
\usepackage{caption} 
\usepackage{amsmath}
\usepackage{algorithm}
\usepackage{algorithmic}
\usepackage{makecell}
\usepackage{subcaption}
\usepackage{color}
\usepackage{booktabs}
\usepackage{multirow}
\usepackage{tcolorbox}
\usepackage{listings}
\usepackage{multirow}
\usepackage[table]{xcolor}
\usepackage{makecell}
\usepackage{tcolorbox}
\usepackage[table]{xcolor} 
\def\revised{\textcolor{blue}}

\def\checklist{\textcolor{black}}
\usepackage{newfloat}
\usepackage{listings}
\usepackage{subcaption} 
\DeclareCaptionStyle{ruled}{labelfont=normalfont,labelsep=colon,strut=off} 
\floatstyle{ruled}
\newfloat{listing}{tb}{lst}{}
\floatname{listing}{Listing}
\title{ProCache: Constraint-Aware Feature Caching with Selective Computation\\for Diffusion Transformer Acceleration}
\author{
    Fanpu Cao\textsuperscript{\rm 1}\equalcontrib, Yaofo Chen\textsuperscript{\rm 1}\equalcontrib\thanks{Corresponding authors.}, Zeng You\textsuperscript{\rm 1}, Wei Luo\textsuperscript{\rm 2,3}
}
\affiliations{
    \textsuperscript{\rm 1}South China University of Technology, China\\
    \textsuperscript{\rm 2}South China Agricultural University, China\\
    \textsuperscript{\rm 3}Pazhou Laboratory, China\\

    chenyaofo@scut.edu.cn
    
}

\usepackage{bibentry}

\begin{document}

\maketitle

\begin{abstract}
Diffusion Transformers (DiTs) have achieved state-of-the-art performance in generative modeling, yet their high computational cost hinders real-time deployment.
While feature caching offers a promising training-free acceleration solution by exploiting temporal redundancy, existing methods suffer from two key limitations: (1) uniform caching intervals fail to align with the non-uniform temporal dynamics of DiT, and (2) naive feature reuse with excessively large caching intervals can lead to severe error accumulation.
In this work, we analyze the evolution of DiT features during denoising and reveal that both feature changes and error propagation are highly time- and depth-varying.
Motivated by this, we propose ProCache, a training-free dynamic feature caching framework that addresses these issues via two core components: 
(i) a constraint-aware caching pattern search module that generates non-uniform activation schedules through offline constrained sampling, tailored to the model’s temporal characteristics; and 
(ii) a selective computation module that selectively compute within deep blocks and high-importance tokens for cached segments to mitigate error accumulation with minimal overhead. 
Extensive experiments on PixArt-$\alpha$ and DiT demonstrate that ProCache achieves up to 1.96$\times$ and 2.90$\times$ acceleration with negligible quality degradation, significantly outperforming prior caching-based methods.
\end{abstract}

\begin{links}
    \link{Code}{https://github.com/macovaseas/ProCache}
\end{links}

\section{Introduction}
Diffusion models have achieved remarkable success in various visual generation tasks, including image and video synthesis~\cite{sd, sora}. Traditional diffusion models predominantly employed CNN-based U-Net architectures~\cite{sd}, while recent transformer-based diffusion models (DiT)~\cite{DiT, sora} have established state-of-the-art performance by scaling up model parameters and training data. However, such superior performance comes at the cost of significant computational overhead, which severely limits the deployment of DiT in real-word applications.

To address this, plenty of efforts have been made to accelerate DiT,
including layer/token pruning \cite{Structural-Pruning, tinydiffusion,sito}, quantization \cite{posttrain, temporaldynamic, hqdit}, and knowledge distillation \cite{snapfusion, fastsampling}.
However, these approaches typically require additional post-training cost and struggle to achieve satisfactory performance under high acceleration ratios.
In contrast, feature caching methods~\cite{deepcache, cancache, fora, teacache} offer a training-free alternative by leveraging temporal redundancy in denoising steps.
They cache computed features for reuse in subsequent steps, significantly improving inference speed without any training requirement. Their plug-and-play nature has attracted substantial interest.

However, these caching-based methods have its natural limitations. First, uniform activation intervals (\textit{e.g.}, full computation every $N$ steps) fail to capture the dynamic characteristics in denoising process.
Although such rigid schedules mitigate error accumulation through periodic updates, they overlook the time-varying nature of feature evolution across steps.
Second, the exponential decay of feature similarity with increasing timestep gaps causes significant error accumulation when reusing stale features. Existing approaches like $\Delta$-DiT~\cite{delta-dit} and FORA~\cite{fora} directly reuse features without refinement, leading to quality degradation at high acceleration ratios.

\begin{figure}[t]
  \centering
  \includegraphics[width=0.47\textwidth]{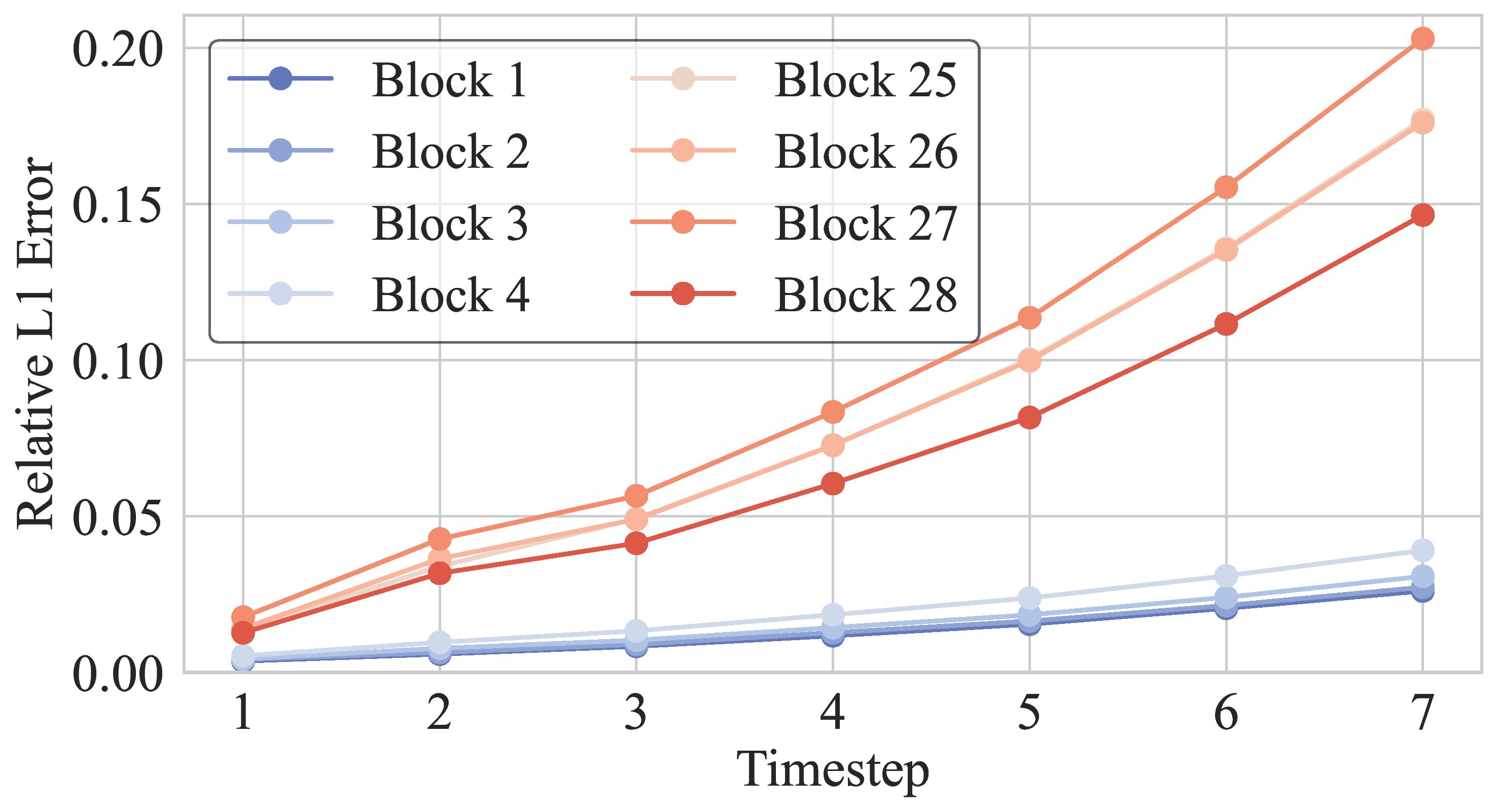}
  \caption{Evolution of relative L1 error across diffusion steps in DiT blocks, which is computed from 10 samples on PixArt-$\alpha$. Errors grow progressively, with deeper blocks (\textit{e.g.}, Block 25–28) showing significantly higher magnitudes than shallower ones (\textit{e.g.}, Block 1–4), highlighting the non-uniform error accumulation across network depths.}
  \label{fig:feature_error}
\end{figure}

To better understand these limitations, we conduct a quantitative analysis of feature error across DiT blocks throughout the denoising trajectory. Our investigation reveals two key observations.
First, as illustrated in Figure~\ref{fig:feature_error}, although feature errors grow progressively over diffusion steps, their accumulation varies significantly across network depths. In particular, deeper blocks exhibit substantially higher error magnitudes.
Second, model outputs exhibit distinct temporal dynamics across denoising steps, with gradual stabilization in the early and middle stages followed by rapid changes (aligning with findings in \cite{smoothecache, temporaldynamic, teacache} and also evident in our empirical studies of Figure~\ref{fig:output_error}). This suggests that a uniform caching schedule fails to align with the evolving nature of the denoising process and may lead to inaccuracies caching reuse in fast-changing phases.
This suggests that a uniform caching strategy, as adopted in prior work, is fundamentally suboptimal.  

Inspired by these, we propose \textbf{ProCache}, a training-free dynamic caching framework for efficient diffusion transformer inference. 
Our ProCache consists of two key components:  
1) Constraint-aware caching pattern search mechanism replaces uniform-interval schedules with offline-optimized, non-uniform activation patterns. By analyzing the evolution of transformer activations across timesteps, we identify phases of high semantic sensitivity and design a constrained sampling procedure to generate caching patterns that maximize reuse in stable stages while enforcing frequent computation during critical one.
2) Selective computation strategy that mitigates error accumulation in long reuse intervals through lightweight, structured updates. Instead of full computation, we selectively refresh only a small set of transformer blocks and tokens. This partial update focuses computational effort where it matters most, effectively maintaining semantic consistency with minimal overhead.

\noindent\textbf{Contributions:} 
1) We analyze that DiT features evolve non-uniformly during inference, \textit{i.e.}, stable in early stages but highly dynamic in later timesteps, with error propagation concentrated in deep layers, revealing the critical mismatch of uniform caching.
2) We propose ProCache, a training-free acceleration framework that solves this via constraint-aware caching pattern search and selective computation, achieving adaptive speedup without quality loss.
3) Experiments show ProCache delivers up to 1.96$\times$ and 2.90$\times$ acceleration on PixArt-$\alpha$ and DiT, respectively, with negligible degradation—outperforming prior methods by a large margin.

\section{Related Work}

\noindent
\textbf{Diffusion models} (DMs) have recently adopted Transformer based architectures, outperforming U-Net based counterparts across multiple domains~\cite{pixart-a, latte, open-sora}, which introduces substantial computational overhead during inference, rendering real-time deployment challenging.
Efforts to mitigate computational demands have explored model compression techniques such as pruning~\cite{Structural-Pruning, token-merging, tinydiffusion, sito}, quantization~\cite{posttrain, temporaldynamic, hqdit}, and knowledge distillation~\cite{fastsampling, snapfusion}. These approaches, however, require resource-intensive retraining or fine-tuning to maintain generation quality.
Consequently, recent research has shifted toward training-free acceleration methods, including \textit{feature caching} and \textit{sampling timestep reduction} methods.

\noindent\textbf{Feature Caching} methods accelerate diffusion models by reusing intermediate features across sampling steps, exploiting temporal redundancies in denoising computations.
Early approaches like DeepCache~\cite{deepcache} and Faster Diffusion~\cite{fasterdiffusion} targeted U-Net architectures, reusing low-resolution features or skipping encoder computations. However, they rely on U-Net's hierarchical structure limits applicability to transformer-based diffusion models.
Recent techniques address this gap through specialized reuse strategies: FORA~\cite{fora} and $\Delta$-DiT~\cite{delta-dit} leverage attention feature reuse and MLP representation sharing, while PAB~\cite{PAB} optimizes attention head computations within DiT blocks. L2C~\cite{L2C} introduces dynamic computation routing through trainable layer-wise skip decisions, achieving higher acceleration ratios at the cost of substantial training overhead. Despite these advances, these methods apply identical caching strategies across all tokens, ignoring fine-grained feature dynamics. To address this, token-level optimization methods like ToCa~\cite{ToCa} selectively caching tokens based on importance scores.

\begin{figure*}[t]
  \centering
  \includegraphics[width=0.98\textwidth]{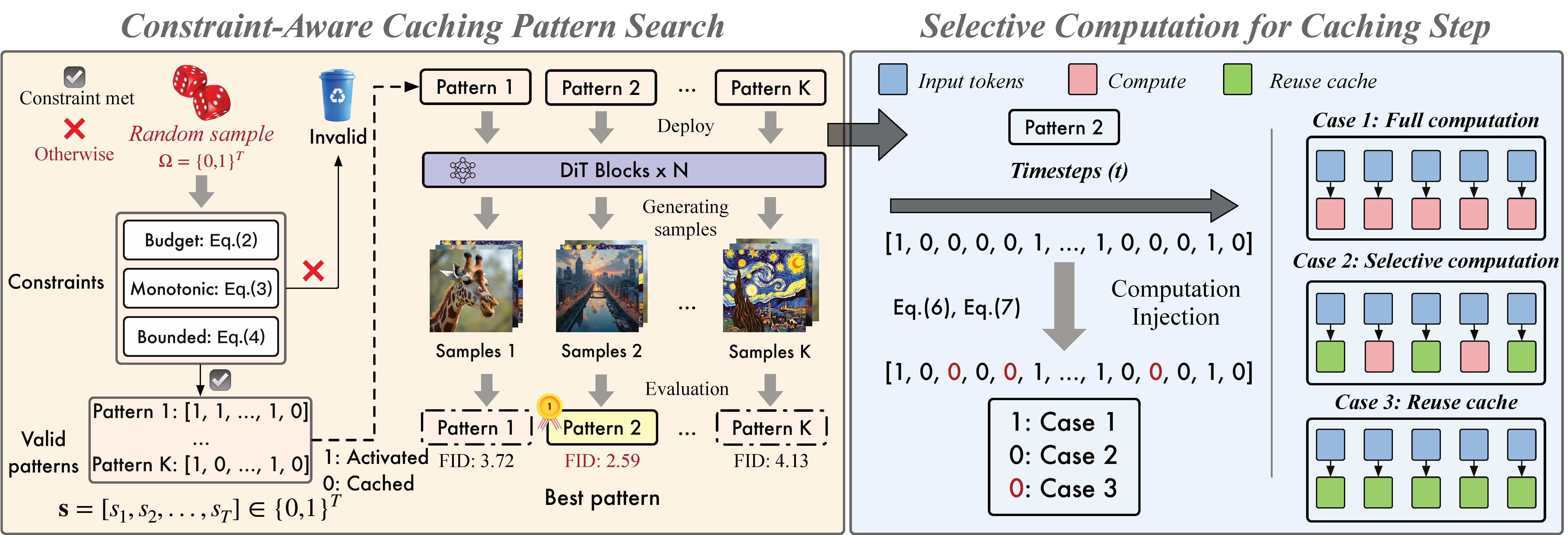}
  \caption{Overall pipeline of ProCache. 1) ProCache explores valid caching patterns under three principled constraints and selects the optimal strategy via lightweight offline sampling based on quality metrics (e.g., FID). 2) It then inserts partial recomputation within contiguous cache steps, selectively updating high-importance tokens in deeper layers.}
  \label{fig:overall}
\end{figure*}

\noindent\textbf{Sampling Timesteps Reduction} methods constitute another prominent approach for accelerating diffusion models, operating by minimizing the number of iterative denoising steps required during generation. While early approaches like DDIM~\cite{DDIM} established deterministic non-Markovian reverse processes for significant step reduction, subsequent advances have refined this paradigm: The DPM-Solver series~\cite{dpm-solver,dpm++} introduced high-order adaptive solvers for optimized differential equation solving, while Rectified Flow~\cite{flowmatching} employs recurrent flow fields to straighten probability paths. Specialized extensions further enhance applicability, YONOS~\cite{yonos} enables single-step super-resolution via scale distillation, DC-Solver~\cite{dc-solver} addresses predictor-corrector misalignment through dynamic compensation, and InvSR~\cite{InvSR} facilitates arbitrary-step sampling via noise predictor networks. Notably, these techniques are orthogonal to the feature caching methods that our work advances.

\section{Proposed Method}

\subsection{Problem Statement and Overview}

Let $\mathcal{G}$ denote a pretrained Diffusion Transformer (DiT)~\cite{DiT} model with $L$ layers, \textit{i.e.}, $\mathcal{G} = g_1 \circ g_2 \circ \cdots \circ g_L$.
Each layer $l$ consists of three components: self-attention (SA), cross-attention (CA), and a multilayer perceptron (MLP), \textit{i.e.}, $g_l = \mathcal{F}^{\text{SA}}_l \circ \mathcal{F}^{\text{CA}}_l \circ \mathcal{F}^{\text{MLP}}_l$.
All these three operations can be unified under a common paradigm: $\mathcal{F}(\mathbf{x}) = \mathbf{x} + \text{AdaLN} \circ f(\mathbf{x})$, where $f(\mathbf{x})$ represents either attention or MLP transformations, and AdaLN~\cite{DiT} ensures stable feature normalization across varying noise conditions. During inference, DiT generates outputs progressively through multiple timesteps from $1$ to $T$. However, this process can be time-consuming due to the sequential computation over many timesteps.

To accelerate inference, we can exploit temporal redundancy by reusing features across timesteps~\cite{fora}. Specifically, instead of recomputing $\mathcal{F}(x_t^l)$ at every timestep $t$, we reuse previously computed features by approximating $\mathcal{F}(\mathbf{x}_{t-k}^l) := \mathcal{F}(\mathbf{x}_t^l)$, where $\mathbf{x}_t^l$ represents the features at timestep $t$ and layer $l$ and $k$ represents the offset in timesteps. 
However, existing approaches typically adopt a uniform interval for feature reuse across all timesteps, failing to account for the non-uniform temporal dynamics inherent in DiT.
Furthermore, these methods propagate cached features across multiple denoising steps without adaptive refinement. As a result, such strategies may lead to suboptimal approximations, limiting acceleration gains due to unnecessary recomputation or inaccurate feature reuse.

\noindent\textbf{Method Overview}.
In this paper, we propose \textit{ProCache}, a training-free caching framework for efficient DiT inference. As illustrated in Figure~\ref{fig:overall}, our ProCache consists of two core components:
1) We propose a constraint-aware caching pattern search method that aims to replace uniform-interval caching pattern with a model-tailored one (c.f. Section~\ref{sec:Ada}). We devise an offline constrained sampling pipeline that searches for promising caching patterns under three principled constraints. The resulting pattern maximizes feature reuse in stable denoising steps while ensuring frequent computation during high-sensitivity ones.
2) We devise a selective computation framework to mitigate error accumulation in long reuse intervals (c.f. Section~\ref{sec:partial_computation}). Instead of fully computing features, we selectively compute only a small subset of transformer blocks and tokens at chosen timesteps.
This effectively refreshing critical semantic content while maintaining low computational overhead.
The overall pipeline of ProCache is shown in Algorithm~\ref{alg:procache_inference}.

\subsection{Constraint-Aware Caching Pattern Search}
\label{sec:Ada}

\begin{figure}[t]
  \centering
  \includegraphics[width=0.46\textwidth]{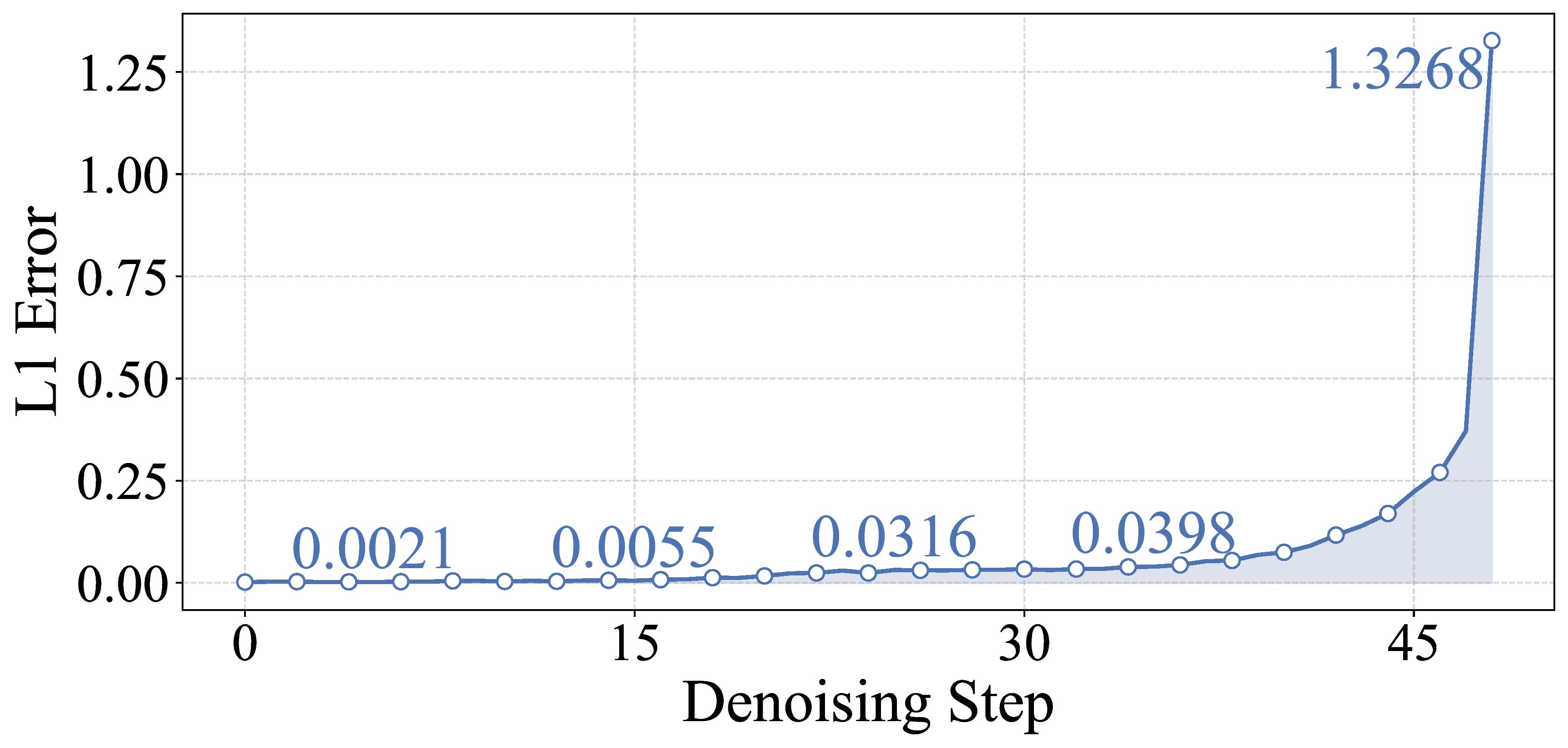}
  \caption{Output L1 error between the current step and the previous step in DiT-XL/2 across the diffusion process.}
  \label{fig:output_error}
\end{figure}

The aforementioned limitations of uniform-interval caching raise a crucial question: \textit{are all timesteps equally conducive to feature reuse?} To investigate this, we measure the temporal discrepancy in DiT-XL-2 outputs via $\|\mathcal{G}_t(\mathbf{x}_t) - \mathcal{G}_{t-1}(\mathbf{x}_{t-1})\|$ across the denoising process. As shown in Figure~\ref{fig:output_error}, feature divergence remains small in early and mid stages but grows sharply in later steps, following an approximately exponential trend.
This reveals that uniform caching misaligns with DiT’s temporal dynamics, causing either excessive computation in stable phases or error accumulation in fast-evolving stages. Instead, effective caching should adapt to the non-uniform evolution of features.

Motivated by this, we propose a constraint-aware caching pattern search framework that finds a customized, non-uniform caching schedule for each model that reduces feature reuse in critical phases and maximizes it where stable.
To this end, we first represents each caching pattern as a binary sequence $\mathbf{s} = [s_1, s_2, \dots, s_T] \in \{0,1\}^T$, where $s_t = 1$ denotes computation at step $t$, and $s_t = 0$ denotes feature reuse from the cache. 
Let $\mathcal{T}(\mathbf{s}) = \{ t \in [T] \mid s_t = 1 \}$ denote the set of timesteps where computation is performed, and let $M = |\mathcal{T}(\mathbf{s})| = \sum_{t=1}^T s_t$ be the total number of activated steps. We define the ordered activation timestamps as $t_1 < t_2 < \cdots < t_M$, obtained by sorting $\mathcal{T}(\mathbf{s})$ in ascending order. We define the $i$-th reuse interval $v_i$ as:
\begin{equation}
v_i = t_{i+1} - t_i - 1, \quad \text{for } i = 1, 2, \dots, M-1,
\end{equation}
which denotes the number of consecutive timesteps with cached features between two computed steps.

\begin{algorithm}[t]
\caption{DiT Inference Pipeline of $l$-th Block with our ProCache. \emph{Note: Before the inference, we have identified an effective caching pattern ${\mathbf{s}^*}$ via the procedure in Sec.~\ref{sec:Ada}}.}
\label{alg:procache_inference}
\begin{algorithmic}[1]
\REQUIRE The DiT block $\mathcal{F}_l$, searched caching pattern ${\mathbf{s}^*}$, selective layer set $\mathcal{U}^{\text{cmpt}}$, selective updated timestep set $\mathcal{T}^{\text{cmpt}}$, token selection ratio $p$, \#denoising steps $T$

\FOR{$t = 1$ \TO $T$}
\IF{$\mathbf{s}^*_t = 1$}
    \STATE \emph{// \textbf{Case 1: Computation without cache}}
    \STATE $\mathcal{F}_l({\bf x}_t) = {\bf x}_t + f_l({\bf x}_t)$; store cache ${\bf c}=f_l({\bf x}_t)$
    \STATE \emph{// AdaLN is omitted for simplicity, same below}
\ELSE
    \IF{$l \in \mathcal{U}^{\text{cmpt}}$ and $t \in \mathcal{T^{\text{cmpt}}}$}
    \STATE \emph{// \textbf{Case 2: Selective computation}}
    \STATE Calculate $\mathcal{I}^{\text{cmpt}}$ via Eqn.(\ref{eq:token_impt}) with $p$
    \STATE $\mathcal{F}_l({\bf x}_t) = {\bf x}_t + f_l({\bf x}_t;{\bf c};\mathcal{I}^{\text{cmpt}})$; where ${\bf c}$ is from cache \emph{// We only compute on important tokens, while the remaining tokens use cache}
    \STATE Store cache ${\bf c}=f_l({\bf x}_t;{\bf c};\mathcal{I}^{\text{cmpt}})$
    \ELSE
    \STATE \emph{// \textbf{Case 3: No computation, use cache}}
    \STATE $\mathcal{F}_l({\bf x}_t) = {\bf x}_t + {\bf c}$; where ${\bf c}$ is from cache
    \ENDIF
\ENDIF
\ENDFOR

\RETURN $\mathcal{F}_i({\bf x}_t)$ \emph{// as the input for next block computation}
\end{algorithmic}
\end{algorithm}

When sampling candidate sequences from the search space $\Omega = \{0,1\}^T$, we hope the candidates satisfied the following three constraints, designed to balance computational efficiency, generation stability, and alignment with DiT's temporal dynamics:

\begin{enumerate}
  \item \textbf{Budget constraint:}  
  To ensure significant acceleration, the total number of activation timesteps must remain below a predefined budget $B \ll T$:
  \begin{equation}\label{eq:budget}
    M = \sum_{t=1}^T s_t \leq B.
  \end{equation}

  \item \textbf{Monotonic constraint:}  
  As shown in Figure~\ref{fig:output_error}, feature changes in DiT are slow in early stages but accelerate in later stages. To preserve fidelity where it matters most, reuse intervals should be non-increasing over time:
  \begin{equation}\label{eq:monotonic}
    v_{i+1} \leq v_i, \quad \forall i = 1,\dots,M-2.
  \end{equation}

  \item \textbf{Bounded constraint:}  
  Excessively long reuse intervals can lead to error accumulation, while overly short ones undermine efficiency. Therefore, each interval is constrained within a feasible range:
  \begin{equation}\label{eq:bounded}
    v^{\min} \leq v_i \leq v^{\max}, \quad \forall i = 1,\dots,M-1,
  \end{equation}
  where $v^{\min} \geq 0$ and $v^{\max} \geq v^{\min}$ are hyper-parameters.
\end{enumerate}

\begin{figure}[t]
  \centering
  \includegraphics[width=0.46\textwidth]{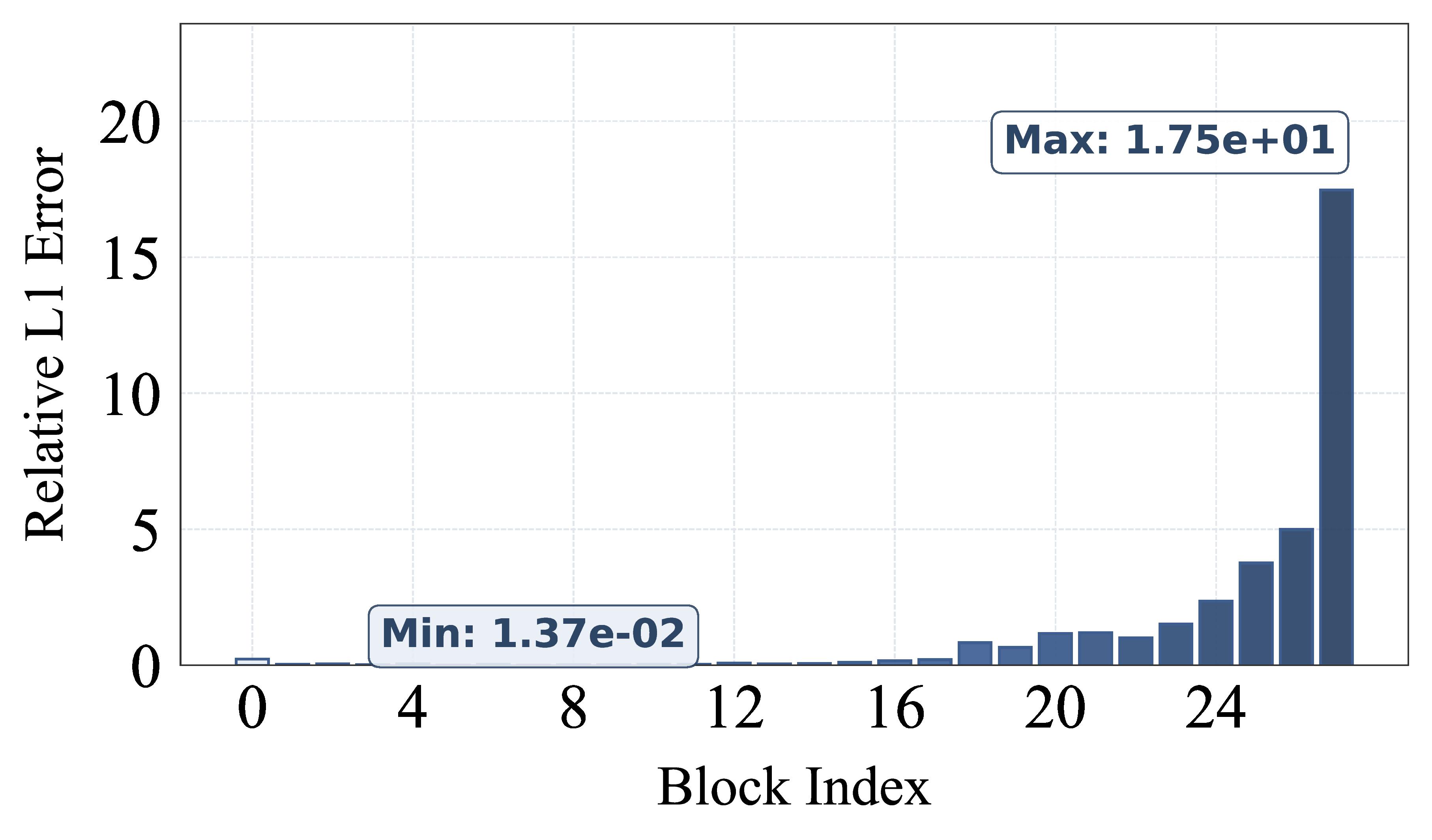}
  \caption{Relative L1 error of features across different blocks in DiT-XL/2, measured at step 20.}
  \label{fig:block_error}
\end{figure}

Formally, we define the constrained search space as the set of all patterns that satisfy all three above constraints:
\begin{equation}
\mathcal{C} {\small{=}} \left\{ \mathbf{s} {\small{\in}} \{0,1\}^T \;\middle|\; \mathbf{s} \text{ satisfies Eqns.} \eqref{eq:budget}, \eqref{eq:monotonic}, \text{ and } \eqref{eq:bounded} \right\}.
\end{equation}
By restricting the search to this constrained space, we efficiently eliminate invalid patterns while preserving effective variations that align with the temporal dynamics of DiT.

\noindent\textbf{Effective Caching Patterns Derivation}.
To find the final caching pattern for a given DiT model, we follow an automated, offline procedure that combines constrained sampling with lightweight evaluation.
We first generate $K$ candidate patterns $\mathbf{s}^{(1)}, \mathbf{s}^{(2)}, \dots, \mathbf{s}^{(K)}$ from the constrained search space $\mathcal{C}$. 
Then, we evaluate each candidate on a small representative dataset on a quality metric such as FID. We adopt the pattern that achieves the best performance as the final pattern $\mathbf{s}^*$.
This process is fully training-free, and completes less than one hour on a single GPU.

\begin{table*}[t]
\centering
\begin{tabular}{l | c c c | c c | c c | c}
\toprule
\bf Method  & \bf Latency(s) $\downarrow$ & \bf FLOPs(T) $\downarrow$ & \bf Speed $\uparrow$  & \bf FID $\downarrow$ & \bf sFID $\downarrow$  & \bf Precision $\uparrow$ & \bf Recall $\uparrow$ &   \makecell{\bf Inception\\ \bf Score} $\uparrow$  \\
\toprule
{\textbf{$\text{DDPM-250 steps}$}} & {49.564}  & {118.68}  & -  &  {{2.31}} &  {{4.98}} & 0.82 & 0.58 & {243.42}\\
\midrule
{\textbf{$\text{DDIM-50 steps}$}} & {4.549}  & {23.74}  & {1.00$\times$}  &  {{2.43}} &  {{4.40}} & 0.80 & 0.59 & {241.25}\\
{\textbf{$\text{DDIM-25 steps}$}} & {2.263}  & {11.87}  & {2.00$\times$}  &  {{3.18}} &  {{4.74}} & 0.79 & 0.58 & {232.01}\\
{\textbf{$\text{DDIM-20 steps}$}} & {1.862}  & {9.49}  & {2.50$\times$}  &  {{3.81}} &  {{5.15}}  & {0.78} &  {0.58} & 221.43 \\
{\textbf{$\text{DDIM-17 steps}$}}  & {1.563}  & {8.07}  & {2.94$\times$}  &  {{4.58}} &  {{5.76}}  & {0.77} &  {0.56} & 208.72\\
\midrule
{\textbf{$\text{$\Delta$-DiT}$}}($\mathcal{N}=3$) & {2.572}  & {16.46}  & {1.47$\times$}  &  {3.75} &  {5.70} & 0.77 & 0.54 &{207.57}\\
{$\textbf{FORA} $} ($\mathcal{N}=3$)  & 2.191 & 8.59 & {2.76$\times$}  & 3.88 & 6.43 & 0.79 & 0.54 & 229.02\\
\textbf{\texttt{\texttt{ToCa}}} ($\mathcal{N}=3$)  & 2.087 & 10.23 & {2.32$\times$}  & 3.04 & 5.14 & 0.79 & 0.56 & 230.70\\
\textbf{\texttt{\texttt{ToCa}}} ($\mathcal{N}=4$)  & 2.063 & 8.73 & {2.72$\times$}  & 3.64 & 5.55 & 0.78 & 0.56 & 223.25\\
{\textbf{ProCache (Ours)}} & \textbf{1.725} & \textbf{8.18} & \textbf{2.90$\times$}  & \textbf{2.96} & \textbf{4.93} & \textbf{0.80} & \textbf{0.57} & \textbf{232.85}\\
\bottomrule
\end{tabular}
\caption{Quantitative comparison on class-to-image generation on ImageNet with DiT-XL/2.}
\label{table:DiT}
\end{table*}

\subsection{Selective Computation for Caching Step}
\label{sec:partial_computation}

Given the searched caching pattern $\mathbf{s}^*$, we incorporate lightweight computations to mitigate error accumulation over multiple denoising steps.
Specifically, we define the set of timesteps for computation based solely on the structure of $\mathbf{s}^*$.
Let $\ell(t)$ denote the starting timestep of the maximal contiguous zero block containing $t$, defined as:
\begin{equation}
\ell(t) = \max \left\{ \tau \leq t \mid s_\tau = 1 \right\} + 1,
\end{equation}
with $\ell(t) = 1$ if no such $\tau$ exists. We apply selective computation at every second position within each zero block. Formally, the set of timesteps selected for updates is:
\begin{equation}
\mathcal{T}^{\text{cmpt}} = \left\{ t \mid s_t = 0,\ t - \ell(t) + 1 \text{ is even} \right\}.
\end{equation}
At each $t \in \mathcal{T}^{\text{cmpt}}$, instead of reusing all cached features, we compute only a subset of layers and tokens, while keeping the rest cached. Formally, the update rule is:
\begin{equation}
\mathbf{x}_i^{(l)} =
\begin{cases}
f^{(l)}(\mathbf{x}_i^{(l-1)}), & \text{if } l \in \mathcal{U}^{\text{cmpt}} \text{ and } i \in \mathcal{I}^{\text{cmpt}} \\
\text{Cache}^{(l)}(\mathbf{x}_i), & \text{otherwise}
\end{cases}
\end{equation}
where $\mathbf{x}_i^{(l)}$ denotes token $\mathbf{x}_i$ at layer $l$, $\mathcal{U}^{\text{cmpt}}$ is the set of layers to compute, $\mathcal{I}^{\text{cmpt}}$ denotes the selected token index, $f^{(l)}(\cdot)$ denotes the computation operator at the $l$-th layer, and $\text{Cache}^{(l)}(\cdot)$ retrieves the cached activation. 
Notably, our selective computation strategy introduces only about 3\% additional latency, as we compute on a small set of caching steps, a subset of layers (25\%), and a subset of tokens (7–30\%).
We detail the construction of $\mathcal{U}^{\text{cmpt}}$ and $\mathcal{I}^{\text{cmpt}}$ below.

\noindent \textbf{Layer Selection: Focus on Deep Blocks}.
To determine the set of layers to compute $\mathcal{U}^{\text{cmpt}}$, we analyze feature error propagation across transformer blocks during the denoising process. As shown in Figure~\ref{fig:block_error}, errors accumulate predominantly in deeper layers, where high-level semantic refinement occurs. While shallow layers exhibit strong temporal stability and minimal deviation over timesteps.
Motivated by this, we define $\mathcal{U}^{\text{cmpt}}$ as the indices of the deepest $D=r*L$ layers: $\mathcal{U}^{\text{cmpt}} = \{ l \mid (L-D) \leq l \leq L \}$, where $r$ is a computed ratio and $L$ is the total number of blocks.

\noindent \textbf{Token Selection: Prioritizing Important Tokens}.  
To determine the set of tokens to compute $\mathcal{I}^{\text{cmpt}}$, we consider their dynamic influence during the generation process. Tokens with large feature changes play a more critical role in shaping the output structure, and thus benefit more from fresh computation. To quantify this, we measure token importance based on the magnitude of its attention module output.
Let ${\bf x}_i$ denote the $i$-th token in the sequence, and let $\mathbf{v}_i$ be its corresponding output from the attention module. We define its importance score as: $T(\mathbf{x}_i) = \|\mathbf{v}_i\|_2$.
We then select the top-$p\%$ most important tokens to form $\mathcal{I}^{\text{cmpt}}$:
\begin{equation}\label{eq:token_impt}
\mathcal{I}^{\text{cmpt}} = \left\{ i \mid \text{rank}(T(\mathbf{x}_i)) \leq p\% \times N \right\},
\end{equation}
where $N$ is the number of tokens. In practice, $p$ is chosen to retain only a small fraction (7–30\%), ensuring minimal computational overhead while preserving key information.

Note that for the cross-attention and FFN modules, we apply this token selection strategy to compute only the top-p\% most important tokens.
However, in the self-attention module, each token attends to all others, making partial computation prone to error propagation.
Thus, we do not employ this token selection strategy for self-attention module.

\section{Experiments}

\subsection{Experiment Setup}

\textbf{Baselines.} We employ three SoTA vision generative models on two tasks: PixArt-$\alpha$ \cite{pixart-a} with 20 DPM-Solver++~\cite{dpm++} steps for text-to-image generation; FLUX.1-dev and FLUX.1-schnell~\cite{flux} utilizing 50 and 4 Rectified Flow \cite{flowmatching} steps respectively for text-to-image generation; and DiT-XL/2~\cite{DiT} with 50 DDIM~\cite{DDIM} steps for class-conditional image generation.

\noindent\textbf{Implementation Details.}
For DiT-XL/2, we perform class-conditional generation on the full 1,000-class ImageNet~\cite{Imagenet}, synthesizing 50,000 images at $256 \times 256$ resolution. PixArt-$\alpha$ text-to-image experiments employed 30,000 randomly selected captions from the COCO-2017~\cite{coco-2017} validation set, generating corresponding images at $256 \times 256$ resolution. For FLUX family, we generated high-resolution $1024 \times 1024$ images on PartiPrompts~\cite{partiprompt} containing 1,632 text prompts.
We set the sampling budget $K$ to 5 and $B \in \{7, 17\}$ in our caching pattern search.
In selective computation, we compute on the deep 25-50\% blocks. $v^{\min}$ and $v^{\max}$ are typically set in the range of 2-5. 
We put more details in the Appendix.

\begin{table*}[t] 
\centering 
\begin{tabular}{ c | l | c | c | c | c } 
\toprule 
{\bf Models} & {\bf Method} & {\bf Latency(s) $\downarrow$} & {\bf FLOPs(T) $\downarrow$} & {\bf Speed $\uparrow$} & \bf Image Reward $\uparrow$ \\ 
\midrule 
\multirow{5}{*}{\parbox{2.5cm}{\centering \textbf{FLUX.1-dev} \\ \citep{flux}}} & {\textbf{$\text{Euler-50 steps}$}} & {33.85} & {3719.50} & {1.00$\times$} & {1.202} \\ 
\cmidrule{2-6}
& {$68\%$\textbf{ steps}} & {23.02} & {2529.26} & {1.47$\times$} & {1.200} \\ 
& $\textbf{FORA}$ \citep{fora} & {20.82} & {2483.32} & {1.51$\times$} & {1.196} \\ 
& {\textbf{\texttt{\texttt{ToCa}} }$(\mathcal{N}=2)$} & 19.88 & 2458.06 & 1.51$\times$ & 1.202 \\ 
& {\textbf{ProCache (Ours)}} & \textbf{18.73} & \textbf{2415.25} & \textbf{1.54$\times$} & \textbf{1.207}\\
\midrule
\multirow{7}{*}{\parbox{2.5cm}{\centering \textbf{FLUX.1-schnell} \\ \citep{flux}}} & {\textbf{$\text{LCM-4 steps}$}} & {2.882} & {277.88} & {1.00$\times$} & {1.133} \\ 
\cmidrule{2-6}
& {$75\%$\textbf{ steps}} & {2.162} & {208.41} & {1.33$\times$} & {1.132} \\ 
& {$\textbf{FORA} ^1$} \citep{fora} & {2.365} & {225.60} & {1.23$\times$} & {1.129} \\ 
& {$\textbf{FORA} ^2$} \citep{fora} & {2.365} & {225.60} & {1.23$\times$} & {1.124} \\ 
& {$\textbf{FORA} ^3$} \citep{fora} & {2.365} & {225.60} & {1.23$\times$} & {1.123} \\ 
& {\textbf{\texttt{\texttt{ToCa}} }$(\mathcal{N}=2)$} & 1.890 & 181.30 & 1.53$\times$ & 1.134\\
& {$\textbf{ProCache (Ours)} (\mathcal{N}=2)$} & \textbf{1.817} & \textbf{177.26} & \textbf{1.56$\times$} & \textbf{1.138}\\
\bottomrule 
\end{tabular} 
\caption{Quantitative comparison in text-to-image generation for FLUX on Image Reward. $\text{FORA}^1$, $\text{FORA}^2$, and $\text{FORA}^3$ represent different step-skipping configurations where the model bypasses step 2, 3, and 4 respectively during generation.}
\label{table:FLUX} 
\end{table*}

\begin{figure*}[t]
  \centering
  \includegraphics[width=0.95\textwidth]{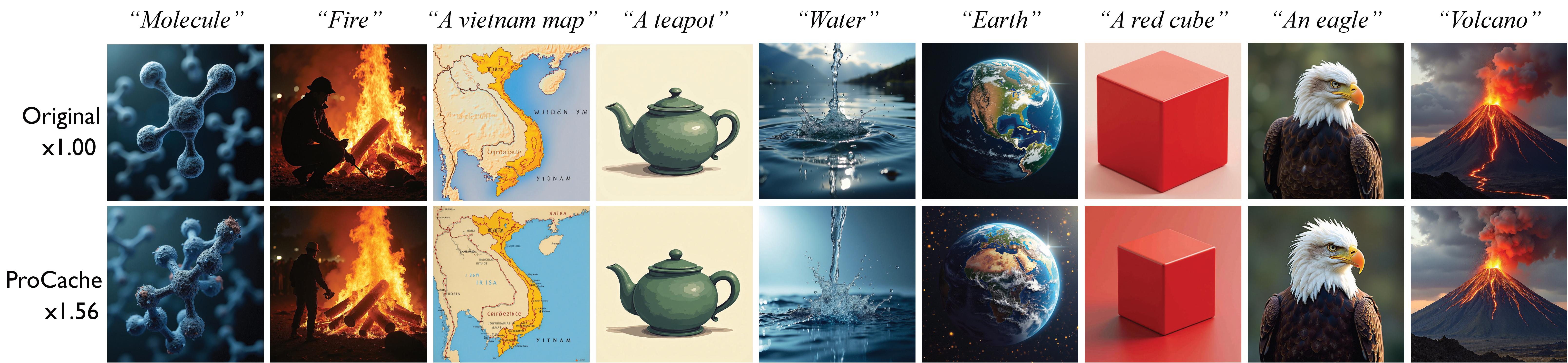}
  \caption{Image generation samples at 1024 $\times$ 1024 resolutions under 1.56$\times$ speed-up ratios.}
  \label{fig:t2img_vis}
\end{figure*}

\subsection{Quantitative Evaluations}
\textbf{Comparisons with SoTA on ImageNet}.
In Table \ref{table:DiT}, our ProCache achieves a 2.90$\times$ speedup while achieving the best FID and sFID scores of 2.96 and 4.93, respectively. Compared to the uniform caching strategy like FORA, our ProCache achieves nearly a 30\% improvement in FID and sFID metrics while enhancing the speedup ratio by 5\%. This result demonstrates that ProCache outperforms other acceleration techniques. Moreover, ProCache maintains competitive Precision, Recall and Inception Score, surpassing existing acceleration methods and reinforcing its advantage.

\begin{table}[t]
    \centering
    \setlength{\tabcolsep}{1mm}
    \fontsize{9pt}{9pt}\selectfont
      \begin{tabular}{l | c c c  | c c }
        \toprule
        \multirow{2}*{\bf Method}  & \multirow{2}*{\bf Latency(s)$\downarrow$} & \multirow{2}*{\bf FLOPs(T)$\downarrow$} & \multirow{2}*{\bf Speed$\uparrow$} & \multicolumn{2}{c}{\bf COCO2017} \\
            & & & & \bf FID$\downarrow$& \bf CLIP$\uparrow$\\
        \midrule
      \textbf{PixArt-$\alpha$} & {2.142} &  11.18  & {1.00$\times$} & 28.12  & {16.29}\\
      \midrule
      {$50\%$\textbf{ steps}}  & {1.044} &  5.59  & {2.00$\times$} & 37.62  & 15.81\\
      {\textbf{$\text{$\Delta$-DiT}$}} & 1.724 &  7.68  & {1.54$\times$} & 28.91  & 16.41 \\
      {$\textbf{FORA}^1$} & 1.575 &  5.66  & {1.98$\times$} & 29.63          & 16.40 \\
      {$\textbf{FORA}^2$} & 1.301 &  6.05  & {2.79$\times$} & 29.84          & 16.42 \\
      \textbf{\texttt{\texttt{ToCa}}} & 1.473 &  6.33  & {1.77$\times$} & 28.02          & 16.43 \\
      {\textbf{ProCache}} & {1.215} &  {5.70} & {1.96$\times$} & \textbf{27.66} & \textbf{16.45}  \\
        \bottomrule
      \end{tabular}
      \caption{Quantitative comparison in text-to-image generation for $\text{PixArt-$\alpha$}$ on MS-COCO2017. $\text{FORA}^1$ and $\text{FORA}^2$ denote configurations with different activation intervals, corresponding to $\mathcal{N}=2$ and $\mathcal{N}=3$, respectively.}
      \label{table:PixArt}
\end{table}

\begin{figure}[t]
  \centering
  \includegraphics[width=0.98\linewidth]{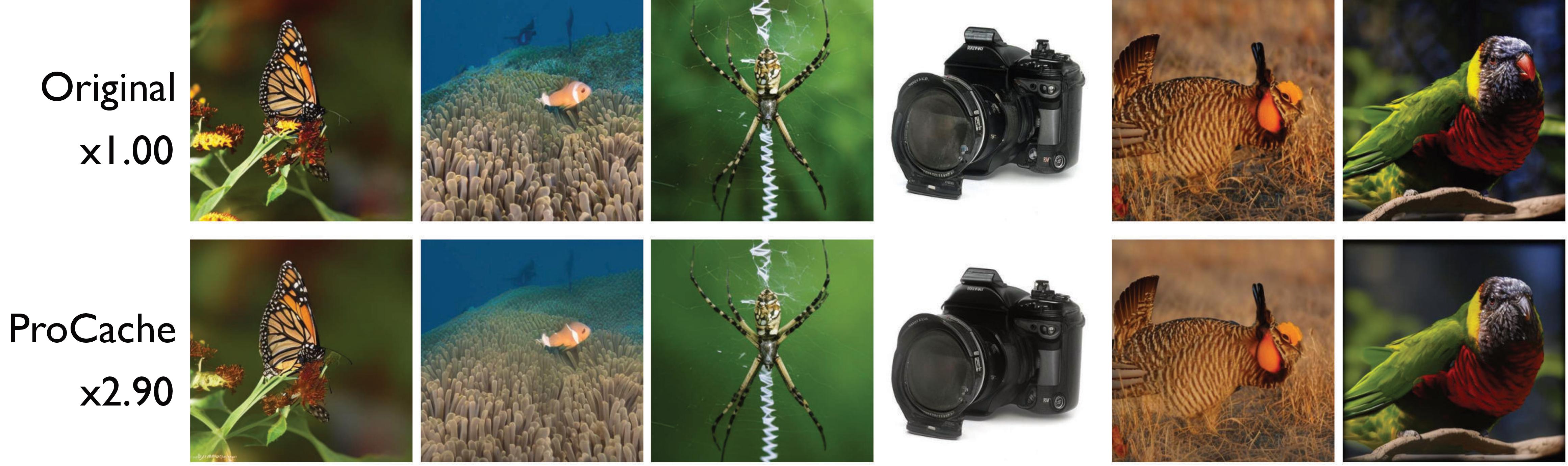}
  \caption{Qualitative comparisons of 512 $\times$ 512 image generations between the original sampler (top, 1.00$\times$ speed) and ProCache (bottom, 2.90$\times$ speed-up). }
  \label{fig:cls_vis}
\end{figure}

\noindent\textbf{Comparisons with SoTA on MS-COCO30k}.
To evaluate ProCache's performance in text-to-image generation tasks, we conducted comprehensive experiments on the COCO30K dataset. As shown in Table~\ref{table:PixArt}, we compare our ProCache with three SoTA training-free acceleration methods across various configuration settings. For comparative analysis, we also included the 10-step DPM-Solver++ sampling approach. 
The quantitative results demonstrate that under an acceleration ratio of approximately 2.0, our ProCache achieves the best trade-off between generation quality and inference speed, outperforming existing methods by securing the lowest FID-30k score and the highest CLIP score on the MS-COCO2017 generation task.

\noindent\textbf{Comparisons with SoTA on PartiPrompts}.
We further evaluate our ProCache on state-of-the-art text-to-image diffusion models featuring higher resolutions and fewer denoising steps, demonstrating its effectiveness under advanced generation conditions. 
Since this model can generate images in just four steps, our ProCache does not use the pattern search strategy and instead adopts a fixed interval of 2 steps for generation. 
In Table \ref{table:FLUX}, our ProCache achieves superior acceleration trade-off compared to existing methods, maintaining comparable Image Reward scores to other baselines at an approximately around 1.55$\times$ speedup ratio on both the 50-step FLUX.1-dev and 4-step FLUX.1-schnell models.
In contrast to FORA that exhibit quality degradation even at 1.5$\times$ acceleration, ProCache preserves perceptual quality metrics while doubling inference efficiency.

\begin{table*}[t]
    \centering
    \subcaptionbox{Ablations on the proposed components.\label{tab:ab_partial}}{%
        \begin{minipage}{0.35\linewidth}
            \centering
            \renewcommand{\arraystretch}{0.98}
            \renewcommand{\tabcolsep}{3pt}
            \begin{small}
            \begin{tabular}{@{}l c c c@{}}
                \toprule
                \bfseries Method & \bfseries Speed$\uparrow$ & \bfseries FID$\downarrow$ & \bfseries sFID$\downarrow$ \\
                \midrule
                Default ($B=13$) & 2.93$\times$ & 4.75 & 8.43 \\
                + Searched Pattern & 2.93$\times$ & 3.15 & 5.12\\
                + Selective Computation & 2.90$\times$ & 3.28 & 5.95\\
                \midrule
                \textbf{ProCache (Ours)} & 2.90$\times$ & \textbf{2.94} & \textbf{4.93}\\
                \bottomrule
            \end{tabular}
            \end{small}
        \end{minipage}
    }\hfill
    \subcaptionbox{Ablations on sample budget $K$.\label{tab:ab_quota}}{%
        \begin{minipage}{0.25\linewidth}
            \centering
            \renewcommand{\arraystretch}{1.35}
            \renewcommand{\tabcolsep}{8pt}
            \begin{small}
            \begin{tabular}{@{}c | c c@{}}
                \toprule
                \bfseries $K$ & \bfseries FID$\downarrow$ & \bfseries Inception Score$\uparrow$ \\
                \midrule
                5 & 45.07 & 181.69 \\
                10 & 45.02 & \textbf{182.50}\\
                15 & \textbf{44.96} & 181.63\\
                \bottomrule
            \end{tabular}
            \end{small}
        \end{minipage}
    }\hfill
    \subcaptionbox{Ablations on proportion $r$ of compute blocks.\label{tab:ab_deep}}{%
        \begin{minipage}{0.35\linewidth}
            \centering
            \renewcommand{\arraystretch}{1.09}
            \renewcommand{\tabcolsep}{3pt}
            \begin{small}
            \begin{tabular}{@{}c|c c c@{}}
                \toprule
                \bfseries $r$ & \bfseries FLOPs (T)$\downarrow$ & \bfseries FID$\downarrow$ & \bfseries Inception Score$\uparrow$ \\
                \midrule
                50\% & 9.138 & 45.32 & \textbf{184.18} \\
                65\% & 8.911 & 45.33 & 184.09 \\
                75\% & 8.594 & \textbf{45.31} & 183.61 \\
                90\% & 8.344 & 45.38 & 182.46 \\
                \bottomrule
            \end{tabular}
            \end{small}
        \end{minipage}
    }
    \caption{Ablation studies of our ProCache on ImageNet with DiT-XL/2.}
    \label{tab:ablation_study}
\end{table*}

\begin{figure*}[t]
    \centering
    \begin{subfigure}[t]{0.33\linewidth}
        \centering
        \includegraphics[width=\linewidth]{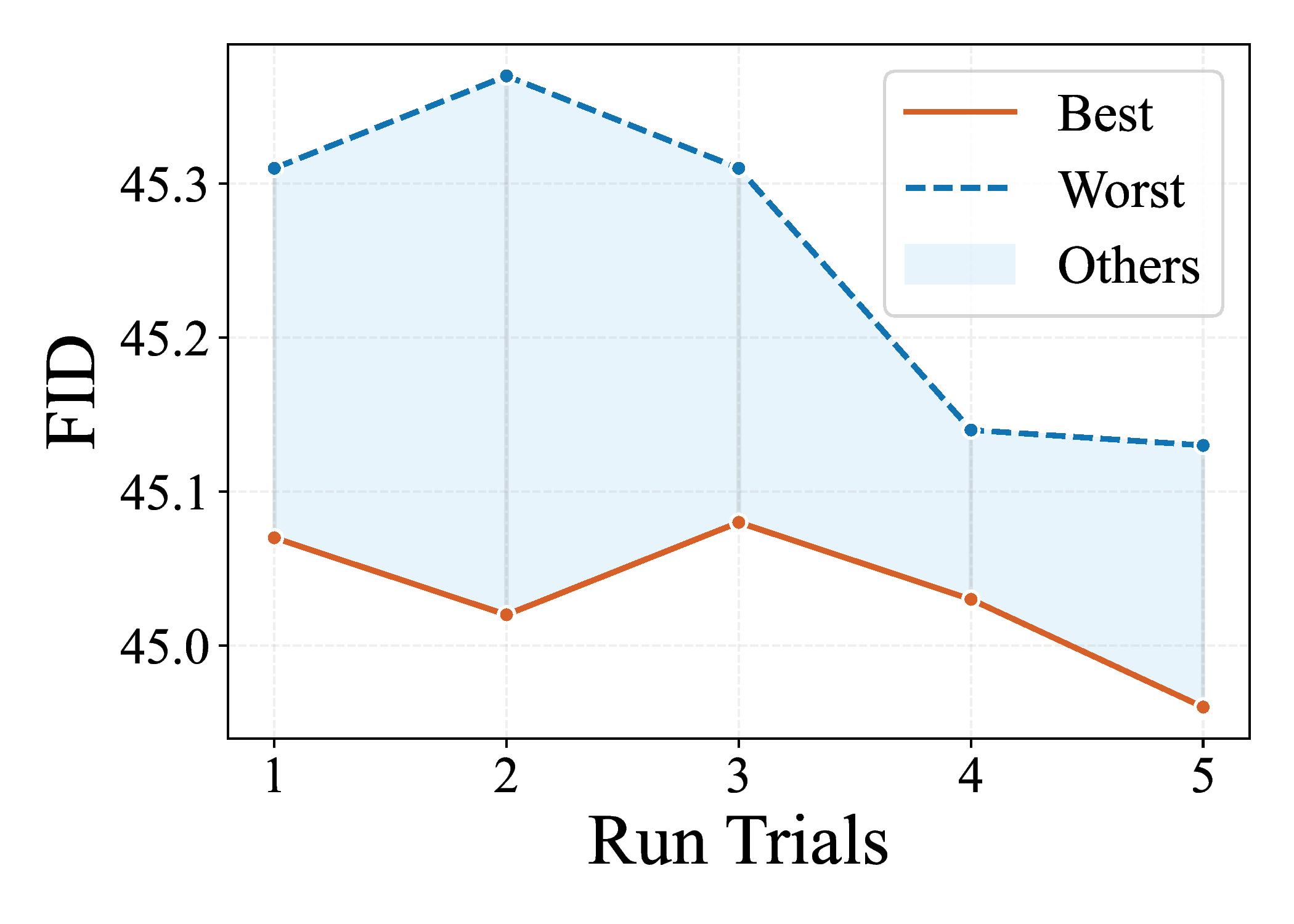}
        \caption{Performance stability across 5 run trials.}
        \label{fig:robustness}
    \end{subfigure}
    \hfill
    \begin{subfigure}[t]{0.63\linewidth}
        \centering
        \includegraphics[width=\linewidth]{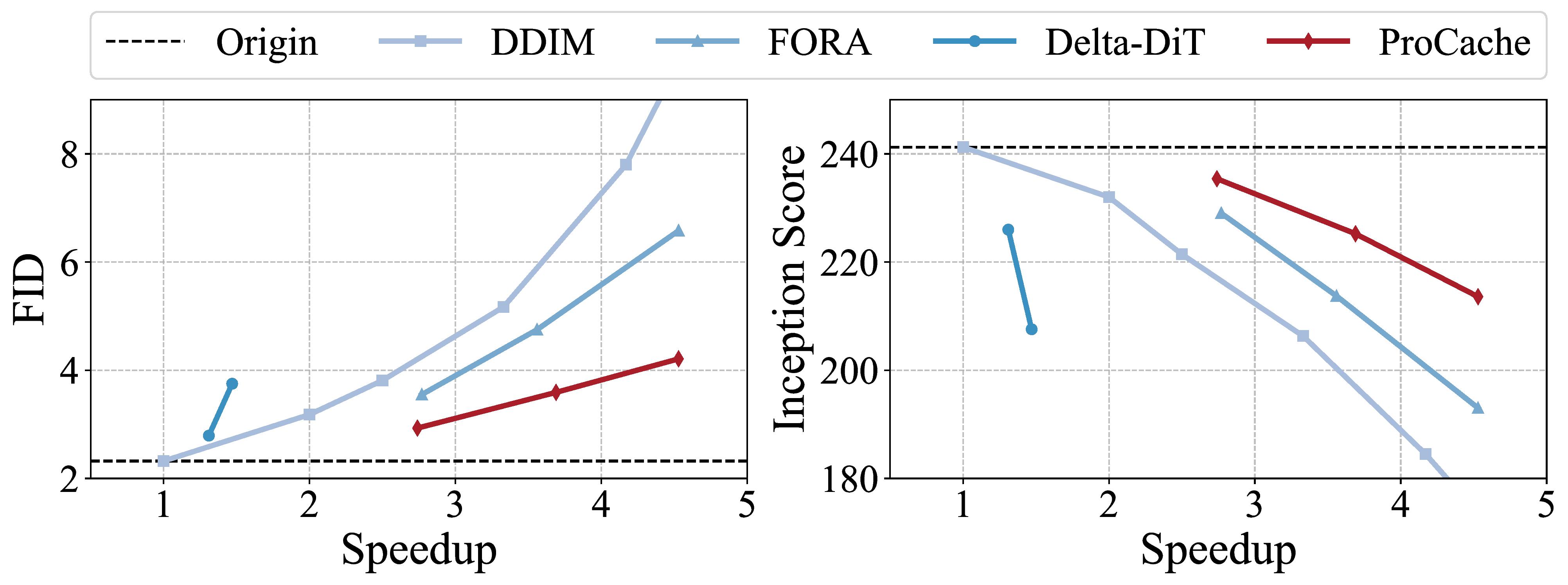}
        \caption{Comparison of speedup ratio versus quality metrics (FID and Inception Score).}
        \label{fig:trade_off}
    \end{subfigure}
    \caption{Robustness and efficiency–quality trade-off analysis of DiT-XL/2 on ImageNet-1k (a) and ImageNet-50k (b).}
    \label{fig:combined}
\end{figure*}

\subsection{Qualitative Analysis}

\noindent \textbf{Text-to-image generation.}
We evaluated ProCache in high-resolution text-to-image generation tasks. As illustrated in Figure \ref{fig:t2img_vis} for FLUX.1-dev, the visual comparison demonstrates ProCache's ability to maintain perceptual quality while achieving substantial acceleration.

\noindent \textbf{Class-conditional image generation.}
In Figure \ref{fig:cls_vis}, we present image generation samples from ProCache. Compared with the original image, ProCache maintains perceptual quality indistinguishable from the original model at 2.90$\times$ acceleration, with imperceptible losses in fine detail.
We put more qualitative results in the Appendix.

\subsection{Further Experiments}

\noindent\textbf{Effect of Different Components}.
In Table~\ref{tab:ab_partial}, the results demonstrates the effectiveness of ProCache's two core components. First, the searched caching pattern improves both FID and sFID without any computational overhead, maintaining identical speed to the default baseline while achieving better generation quality. Second, the selective computation mechanism further enhances performance with minimal additional cost, the slight reduction in speed (2.90$\times$ vs. 2.93$\times$), yet yields measurable improvements. 

\noindent\textbf{Ablations on Sampling Budget}.
In Table~\ref{tab:ab_quota}, we present ablations on sampling budget $K$. As $K$ increases from 5 to 15, FID scores exhibit an improvement, indicating that larger sampling budget enable discovery of higher-quality activation schedules. 
Notably, even with $K=5$, the lowest sampling budget and the least performant setting, our method still surpasses existing approaches.
Given its superior computational efficiency, we adopt $K=5$ as the default budget.

\noindent\textbf{Effect of computing \#blocks}.
To study this, we evaluate performance at different depths (50–90\% of total blocks) on DiT-XL/2.
In Table~\ref{tab:ab_deep}, 75\% of blocks achieves the lowest FID and a competitive Inception Score. Going deeper (\textit{e.g.}, 50\%) brings no FID improvement and lowers Inception Score, indicating diminishing returns beyond a threshold. To maximize speedup, we use 25\% in main experiments—the minimal depth that preserves quality.

\noindent \textbf{Robustness of Caching Pattern Search}.
Our constraint-aware cache pattern search in Sec.~\ref{sec:Ada} involves randomness. To verify its robustness in practical use, we conducted five run trials on the ImageNet-1K. In Figure \ref{fig:robustness}, the variations in both FID and Inception Score across runs are minimal. This indicates that our algorithm always finds promising cache pattern, maintaining reliable generative performance regardless of the random sampling variation.

\noindent \textbf{High-acceleration Ratios Performance}.
To comprehensively evaluate the trade-off between speedup and model accuracy, we conducted experiments on ImageNet-50k using DDIM with varying steps, $\Delta$-DiT, and FORA. In Figure~\ref{fig:trade_off}, our ProCache outperforms prior caching methods, which suffer from feature error accumulation. Our ProCache reduces quality degradation by 56.2\% and achieves robust performance in acceleration regimes exceeding 4.53$\times$.

\section{Conclusion}
In this work, we addressed the challenge of high computational costs associated with DiTs in generative modeling by proposed ProCache, a training-free dynamic feature caching framework. ProCache achieves 2.90$\times$ speedup with minimal impact on generation quality, outperforming existing methods. Our results highlight its potential to enable efficient real-time deployment of DiTs.

\section*{Acknowledgments}
This work was partially supported by Postdoctoral Fellowship Program of CPSF (GZC20251043) and the Young Scholar Project of Pazhou Lab (No.PZL2021KF0021).

\bibliography{aaai2026}

\clearpage


\onecolumn 
\appendix 

\setcounter{figure}{0}
\setcounter{table}{0}
\setcounter{equation}{0}
\setcounter{algorithm}{0}

\renewcommand\thesection{\Alph{section}}
\renewcommand\thefigure{\Alph{figure}}
\renewcommand\thetable{\Alph{table}}
\renewcommand\thealgorithm{\Alph{algorithm}}
\renewcommand\theequation{\Alph{equation}}

\def\revised{\textcolor{blue}}
\def\checklist{\textcolor{black}}

\lstset{%
	basicstyle={\footnotesize\ttfamily},
	numbers=left,numberstyle=\footnotesize,xleftmargin=2em,
	aboveskip=0pt,belowskip=0pt,
	showstringspaces=false,tabsize=2,breaklines=true}

\begin{center}
\LARGE \textbf{Appendix of ``ProCache: Constraint-Aware Feature Caching with Selective Computation for Diffusion Transformer Acceleration''}
\end{center}
\vspace{1cm}


In the supplementary, we provide more experimental results and more details on the ProCache method.
We organize our supplementary as follows.

\begin{itemize}
	\item In Section~\ref{supp:Implement details}, we provide more implementation details for evaluation and model configuration.
    \item In Section~\ref{supp:Method Details}, we provide more details about the proposed ProCache.
    \item In Section~\ref{supp:More results}, we provide additional experimental results.
	\item In Section~\ref{supp:visualization results}, we show more visualization results for qualitative analysis.
    \item In Section~\ref{supp:Final Pattern Details}, we provide the final pattern searched by ProCache for different models.
\end{itemize}

\section{More Implementation Details}
\label{supp:Implement details}

\subsection{More Details of Evaluation}
For DiT-XL/2, we performed class-conditional generation on the full 1,000-class ImageNet dataset~\cite{Imagenet}, synthesizing 50,000 images at $256 \times 256$ resolution. Model optimization was guided by FID-1k~\cite{fid} scores, with final performance evaluated using FID-50k. Secondary evaluation metrics included sFID, Inception Score, and Precision \& Recall metrics. The PixArt-$\alpha$ text-to-image experiments employed 30,000 randomly selected captions from the COCO-2017~\cite{coco-2017} validation set, generating corresponding images at $256 \times 256$ resolution. We conducted parameter optimization using FID-3k and performed final quality assessment with FID-30k. Image-text alignment was quantified using CLIP Score~\cite{clip-score} between generated images and input prompts. For the FLUX family, we generated high-resolution $1024 \times 1024$ images on the PartiPrompts dataset~\cite{partiprompt} containing 1,632 text prompts. We use Image Reward~\cite{imagereward}, a metric that better suited to measuring human preference, to evaluate the generation quality.
All experiments were implemented in Python 3.9 with PyTorch 2.4.0. Computational resources included two NVIDIA RTX 3090 (24GB) GPUs for DiT-XL/2 and PixArt-$\alpha$ experiments, and a single NVIDIA A800 (80GB) GPU for FLUX models evaluations.

\subsection{More Details of Hyper-parameters in Different Diffusion Models}

The ProCache framework employs six key hyper-parameters that govern its dynamic caching and selective computation strategies: $K$ denotes the sampling budget for candidate pattern generation, $B$ represents the activation budget constraining total computational steps, $p$ specifies the token selection ratio for importance-based token computation, $v^{\min}$ and $v^{\max}$ define the minimum and maximum reuse intervals for feature caching, and $r$ indicates the proportion of transformer blocks requiring computation updates. These hyper-parameters collectively balance acceleration efficiency with generation quality by adapting to the non-uniform temporal dynamics of diffusion transformers, with their optimal configurations varying across different model architectures as detailed in Table \ref{tab:hyper-parameter}.

\begin{table}[htbp]
\centering
\begin{tabular}{c|c|c|c|c|c|c}
\toprule
\textbf{Model} & \textbf{$K$} & \textbf{$B$} & \textbf{$p$} & \textbf{$v^{\min}$} & \textbf{$v^{\max}$} & \textbf{$r$} \\
\midrule
DiT-XL/2 & 5 & 17 & 7\% & 2 & 5 & 75\% \\
PixArt-$\alpha$ & 5 & 7 & 30\% & 2 & 3 & 50\% \\
FLUX.1-dev & 5 & 17 & 30\% & 2 & 3 & 50\% \\
FLUX.1-schnell & - & - & 30\% & - & - & 50\% \\
\bottomrule
\end{tabular}
\caption{Hyperparameter settings on different models. In FLUX.1-schnell, we abandoned the pattern search strategy.}
\label{tab:hyper-parameter}
\end{table}

\section{More Details of our ProCache}
\label{supp:Method Details}

\subsection{Constraint-Aware Caching Pattern Search}

The constraint-aware caching pattern search framework lies at the core of ProCache, enabling the discovery of optimal non-uniform caching schedules that align with the temporal dynamics of diffusion transformers. As formally presented in Algorithm~\ref{alg:cass}, our framework employs a constrained random sampling strategy to efficiently explore the vast space of possible caching patterns while respecting three critical constraints identified in Section 3.2 of the main paper.

The algorithm begins by initializing an empty candidate set $\mathcal{C}$ and iteratively generates random binary sequences $\mathbf{s} \in \{0,1\}^T$ where each element $s_t$ indicates whether computation (1) or caching (0) occurs at timestep $t$. For each candidate sequence, the algorithm verifies three essential constraints in a specific order of computational efficiency. First, it checks the \textbf{budget constraint} by ensuring the total number of activation steps satisfies $\sum_{t=1}^{T} s_t \leq B$, discarding sequences that violate this fundamental requirement for target acceleration ratio. Second, the algorithm extracts the sorted activation indices $\mathcal{A} = \{t_1, t_2, \ldots, t_M\}$ (where $t_1 < t_2 < \cdots < t_M$) and verifies the \textbf{bounded constraint} by confirming all inter-activation intervals $v_i = t_{i+1} - t_i - 1$ fall within the specified range $[v^{\min}, v^{\max}]$, which prevents both excessive error accumulation from long reuse intervals and unnecessary computation from overly short ones.

Notably, while the \textbf{monotonic constraint} ($v_{i+1} \leq v_i$) is described in the main paper as ensuring longer reuse intervals in early stable stages and shorter ones in later dynamic stages, our implementation handles this constraint implicitly through the evaluation phase rather than the sampling phase. This design choice significantly improves sampling efficiency while still capturing the desired temporal characteristics—candidate patterns violating the monotonic principle typically receive lower quality scores during evaluation and are thus naturally filtered out.

The algorithm continues this constrained sampling process until either $K$ valid patterns are collected or the maximum attempt count $M$ is reached, ensuring computational tractability. The resulting candidate set $\mathcal{C}$ is then evaluated on a small representative dataset using quality metrics such as FID, with the top-$K$ performing patterns selected for final deployment. This two-stage approach—constrained sampling followed by lightweight evaluation—strikes an optimal balance between exploration efficiency and solution quality, requiring only approximately one hour on a single GPU to identify the optimal caching pattern for a given diffusion transformer model.

\begin{algorithm}[t]
\caption{Constraint-Aware Caching Pattern Search}
\label{alg:cass}
\begin{algorithmic}
\REQUIRE
\begin{tabular}{ll}
$T$ & Total denoising steps \\
$K$ & Sampling quota \\
$B$ & Maximum number of activation steps \\
$v^{\min}$ & Minimum reuse interval \\
$v^{\max}$ & Maximum reuse interval \\
$M$ & Maximum number of attempts
\end{tabular}

\ENSURE
$\mathcal{S} \subseteq \{0,1\}^T$: Set of $K$ feasible cache-activation scheduling strategies

\STATE Initialize an empty set $\mathcal{C} \leftarrow \emptyset$
\STATE Set attempt counter $m \leftarrow 0$

\WHILE{$|\mathcal{C}| < K$ and $m < M$}
    \STATE Generate a random binary sequence $\mathbf{s} \in \{0,1\}^T$ via uniform stochastic sampling
    \STATE $m \leftarrow m + 1$

    \IF{$\sum_{t=1}^{T} s_t > B$}
        \STATE Discard $\mathbf{s}$ due to budget violation
        \STATE Continue to next iteration
    \ENDIF

    \STATE Extract activation indices $\mathcal{A} = \{ t \in [1,T] \mid s_t = 1 \}$, sorted in increasing order

    \FOR{$i = 1$ to $|\mathcal{A}| - 1$}
        \STATE Compute inter-activation interval $v_i = \mathcal{A}_{i+1} - \mathcal{A}_i - 1$
        \IF{$v_i < v^{\min}$ or $v_i > v^{\max}$}
            \STATE Discard $\mathbf{s}$ due to interval constraint violation
            \STATE Break and continue to next iteration
        \ENDIF
    \ENDFOR

    \STATE Add $\mathbf{s}$ to candidate set: $\mathcal{C} \leftarrow \mathcal{C} \cup \{\mathbf{s}\}$
\ENDWHILE

\STATE \textbf{return} $\mathcal{S} \leftarrow$ top-$K$ distinct strategies from $\mathcal{C}$
\end{algorithmic}
\end{algorithm}

\subsection{Selective Computation for Caching Step}
\textbf{Token Selection and Updates.}
DiT architectures allow processing a variable number of tokens, offering flexibility for efficient inference. Leveraging this property, we first identify a critical subset of tokens—referred to as \textit{base tokens}—and restrict heavy computation to only these tokens, while skipping the rest using cached values from previous timesteps. The cache is then selectively updated across layers.

A key challenge lies in identifying these base tokens efficiently. Prior works~\cite{sito, ToCa} often rely on attention-based token importance scores, e.g., by selecting tokens with high attention weights in self-attention maps. However, such strategies require access to attention scores, which are not available in optimized attention implementations like FlashAttention~\cite{flash-attention} and memory-efficient attention. These implementations trade off intermediate score outputs for speed and memory efficiency, making score-based selection incompatible in practice. To resolve this incompatibility, we adopt a score-free yet effective proxy for token importance—namely, the $\ell_2$-norm of the value vector in attention layers. Empirically, tokens with higher attention scores often correspond to larger value vector norms. Hence, the norm of the value projection serves as a practical and efficient criterion for importance estimation. Therefore, we define the importance score of token $x_i$ using its corresponding value vector in the attention module: $\mathbf{v}_i = \text{ValueProj}(x_i)$.

\noindent \textbf{Computation Injection for Long Cache Sequences.}
We consider the risk of prolonged cache reuse. Even with deep-layer updates, long contiguous sequences of $s_t = 0$ can lead to irreversible error drift. To prevent this, we inject lightweight partial computations into maximal contiguous zero blocks in $\mathbf{s}$.

Specifically, for every such block $[0, 0, \ldots, 0]$, we insert partial updates at regular intervals—specifically, at every second position starting from the second:
\begin{align*}
[0] &\rightarrow [0] \\
[0,0] &\rightarrow [0,\alpha] \\
[0,0,0] &\rightarrow [0,\alpha,0] \\
[0,0,0,0] &\rightarrow [0,\alpha,0,\alpha]\\
[0,0,0,0,0] &\rightarrow [0,\alpha,0,\alpha,0]\\
[0,0,0,0,0,0] &\rightarrow [0,\alpha,0,\alpha,0,\alpha]\\
\end{align*}
where $\alpha$ denotes a timestep with partial computation (deep-layer update only). This injection ensures that no cache-only sequence exceeds two consecutive steps without correction, improving stability with minimal overhead.

\section{More Experimental Results}
\label{supp:More results}

\subsection{Demonstration of ProCache's Flexibility}

The flexibility of ProCache is demonstrated in Table \ref{table:DiT_more_results}, which presents comprehensive performance across varying computational budgets. Our framework exhibits remarkable adaptability by allowing precise control over the acceleration-quality trade-off through the activation budget parameter $B$. When configured with $B=21$, ProCache achieves a 2.74$\times$ speedup while delivering state-of-the-art FID, significantly outperforming existing methods at comparable acceleration ratios. With a more aggressive budget of $B=17$, the framework maintains exceptional generation quality while achieving a 2.90$\times$ speedup, surpassing all baseline methods in both quantitative metrics and visual fidelity. Most notably, when pushed to an extreme budget of $B=13$, ProCache still delivers a substantial 3.69$\times$ acceleration with acceptable quality degradation, demonstrating robust performance even in high-acceleration regimes. This tunable characteristic makes ProCache particularly valuable for diverse deployment scenarios—from resource-constrained edge devices requiring maximum acceleration to quality-sensitive applications where minimal degradation is paramount—without requiring any model retraining or architectural modifications.

\begin{table*}[ht]
\centering
\resizebox{\textwidth}{!}{
\begin{tabular}{l | c c c | c c | c c | c}
\toprule
\bf Method & \bf Latency(s) $\downarrow$ & \bf FLOPs(T) $\downarrow$ & \bf Speed $\uparrow$ & \bf FID $\downarrow$ & \bf sFID $\downarrow$ & \bf Precision $\uparrow$ & \bf Recall $\uparrow$ & \makecell{\bf Inception\\ \bf Score} $\uparrow$ \\
\toprule
{\textbf{$\text{DDPM-250 steps}$}} & {49.564} & {118.68} & - & {{2.31}} &{{4.98}} & 0.82 & 0.58 & {243.42}\\
\midrule
{\textbf{$\text{DDIM-50 steps}$}} & {4.549} & {23.74} & {1.00$\times$} & {{2.43}} & {{4.40}} & 0.80 & 0.59 & {241.25}\\
{\textbf{$\text{DDIM-25 steps}$}} & {2.263} & {11.87} & {2.00$\times$} & {{3.18}} & {{4.74}} & 0.79 & 0.58 & {232.01}\\
{\textbf{$\text{DDIM-20 steps}$}} & {1.862} & {9.49} & {2.50$\times$} & {{3.81}} & {{5.15}} & {0.78} & {0.58} & 221.43 \\
{\textbf{$\text{DDIM-17 steps}$}} & {1.563} & {8.07} & {2.94$\times$} & {{4.58}} & {{5.76}} & {0.77} & {0.56} & 208.72\\
\midrule
{\textbf{$\text{$\Delta$-DiT}$}}($\mathcal{N}=3$) & {2.572} & {16.46} & {1.47$\times$} & {3.75} & {5.70} & 0.77 & 0.54 &{207.57}\\
{$\textbf{FORA} $} ($\mathcal{N}=3$) & 2.191 & 8.59 & {2.76$\times$} & 3.88 & 6.43 & 0.79 & 0.54 & 229.02\\
\textbf{\texttt{\texttt{ToCa}}} ($\mathcal{N}=3$) & 2.087 & 10.23 & {2.32$\times$} & 3.04 & 5.14 & 0.79 & 0.56 & 230.70\\
\textbf{\texttt{\texttt{ToCa}}} ($\mathcal{N}=4$) & 2.063 & 8.73 & {2.72$\times$} & 3.64 & 5.55 & 0.78 & 0.56 & 223.25\\
\rowcolor{gray!15}
{\textbf{ProCache}} ($B=21$) & 2.049 & 8.66 & {2.74$\times$} & \textbf{2.93} & \underline{4.94} & \textbf{0.81} & \textbf{0.57} & \textbf{235.40}\\
\rowcolor{gray!15}
{\textbf{ProCache}} ($B=17$) & 1.725 & 8.18 & 2.90$\times$ & \underline{2.96} & \textbf{4.93} & \underline{0.80} & \textbf{0.57} & \underline{232.85}\\
\rowcolor{gray!15}
{\textbf{ProCache}} ($B=13$) & \textbf{1.654} & \textbf{6.42} & \textbf{3.69$\times$} & 3.59 & 6.05 & \underline{0.80} & 0.55 & 225.23\\
\bottomrule
\end{tabular}
}
\caption{Quantitative comparison on class-to-image generation on ImageNet with DiT-XL/2.}
\label{table:DiT_more_results}
\end{table*}

\subsection{Feasibility of Offline Evaluation}

For the ProCache framework, the optimal caching pattern is selected through offline evaluation on a small representative dataset to minimize computational overhead. To validate the feasibility of this approach, we conduct ablation studies on ImageNet where we fix a computational budget and generate 5 random patterns. As shown in Table \ref{tab:offline}, the pattern achieving the best performance on the small dataset (FID-1k) consistently demonstrates superior performance on the full dataset (FID-50k) as well, confirming that offline evaluation on a small subset reliably identifies patterns that generalize to the complete dataset. This critical finding validates our practical design choice of using lightweight offline evaluation, which significantly reduces the search time from hours to minutes while maintaining selection accuracy and eliminating the need for resource-intensive full-dataset evaluation during the pattern search phase.

\begin{table}[htbp]
\centering
\begin{tabular}{c|c c}
\toprule
\textbf{Run Trials} & \bf FID-1k $\downarrow$ & \bf FID-50k $\downarrow$\\
\midrule
1 & 45.44 & 3.41\\
2 & 45.48 & 3.42\\
3 & 45.64 & 3.61\\
4 & 45.45 & 3.46\\
5 & \textbf{45.36} & \textbf{3.39}\\
\bottomrule
\end{tabular}
\caption{Consistency between small-scale (FID-1k) and full-scale (FID-50k) evaluation of caching patterns on ImageNet. The pattern achieving optimal performance on the small dataset consistently generalizes to the complete dataset.}
\label{tab:offline}
\end{table}

\subsection{Computational Efficiency of Pattern Search}

The computational efficiency of our constraint-aware pattern search algorithm is critical for practical deployment. We evaluate its performance on a standard CPU environment to demonstrate its lightweight nature. When configured with the default sampling quota $K=5$, the algorithm completes the search process in approximately 0.001 seconds on an AMD Ryzen 7 5800H processor (3.2GHz). This near-instantaneous execution time confirms that the pattern search introduces negligible overhead to the overall inference pipeline. As expected, the search duration scales linearly with the maximum attempt count $M$, following the relationship $t \propto M$, while remaining well below one second even for significantly larger $M$ values.

To further investigate the relationship between search effort and solution space exploration, we conduct experiments with varying maximum attempt counts while fixing other parameters ($B=17$, denoising steps $T=50$, $v^{\min}=2$, $v^{\max}=5$). Table \ref{tab:search_efficiency} presents the number of valid patterns discovered across different $M$ values, with the sampling quota set to $K=1000$ to maximize solution discovery without artificial constraints. Notably, the results reveal that the constrained search space contains only 30 valid patterns for this configuration. The search process rapidly converges, finding 20 patterns with just $10^3$ attempts and reaching the complete set of 30 patterns by $10^6$ attempts. Crucially, increasing $M$ beyond $10^6$ yields no additional valid patterns, as evidenced by identical counts at $M=10^6$ and $M=10^7$. This convergence behavior confirms that our default setting of $M=10^6$ provides complete exploration of the feasible solution space while maintaining computational efficiency, making higher attempt counts unnecessary for practical implementation.

\begin{table}[htbp]
\centering
\begin{tabular}{c|c|c}
\toprule
\textbf{Maximum Attempts ($M$)} & \textbf{Valid Patterns Found} & \textbf{Search Time (s)} \\
\midrule
$10^3$ & 20 & 0.008 \\
$10^4$ & 25 & 0.086 \\
$10^5$ & 27 & 0.856 \\
$10^6$ & 30 & 8.572 \\
$10^7$ & 30 & 86.07 \\
\bottomrule
\end{tabular}
\caption{Relationship between maximum attempt count and the number of valid patterns discovered under fixed constraints ($B=17$, $T=50$, $v^{\min}=2$, $v^{\max}=5$).}
\label{tab:search_efficiency}
\end{table}

\begin{figure}[h]
\centering
\includegraphics[width=0.85\textwidth]{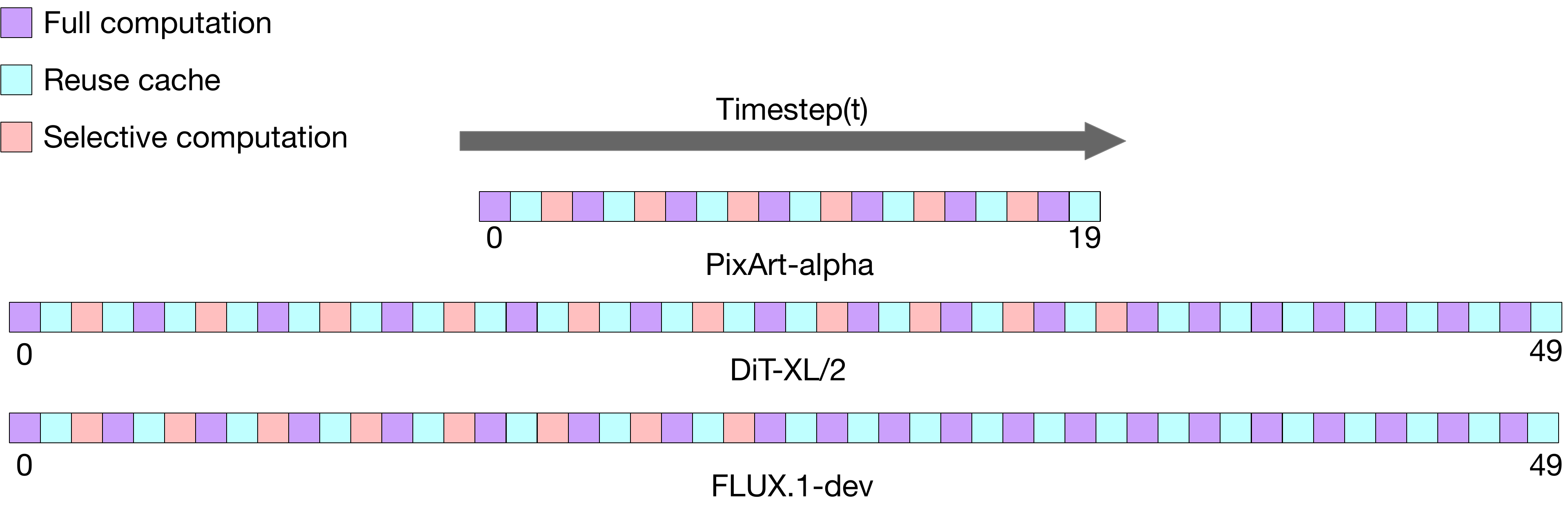}
\caption{Visualization of the final caching patterns searched by ProCache for different models.}
\label{fig:final_patterns}
\end{figure}

\section{More Visualization Results}
\label{supp:visualization results}
\begin{figure}[ht]
\centering
\begin{minipage}{\textwidth}
\centering
\includegraphics[width=0.95\textwidth]{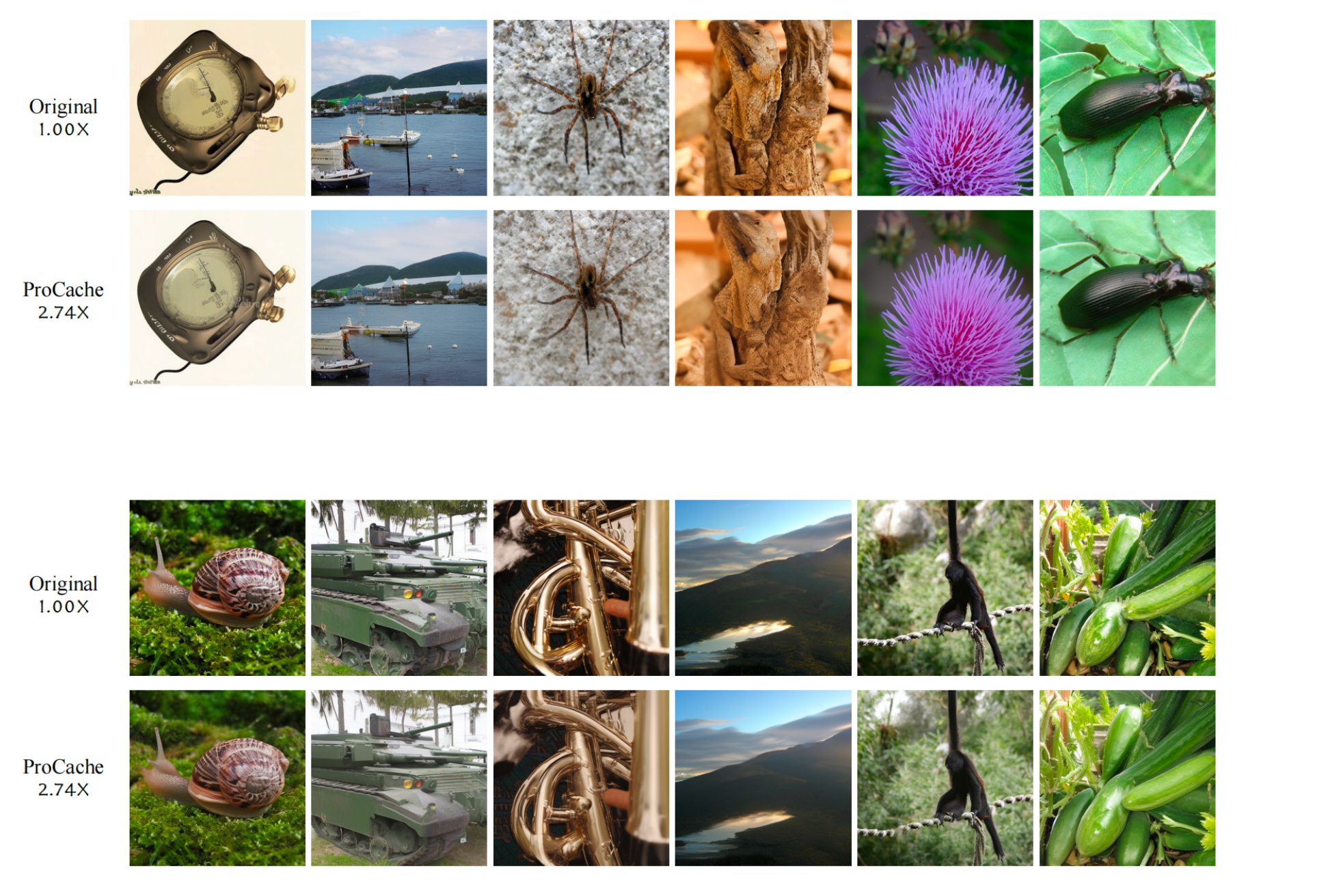}
\captionof{figure}{Visualization examples of DiT-XL/2.}
\label{fig:dit_apx}
\end{minipage}

\vspace{1.5em}

\begin{minipage}{\textwidth}
\centering
\includegraphics[width=0.95\textwidth]{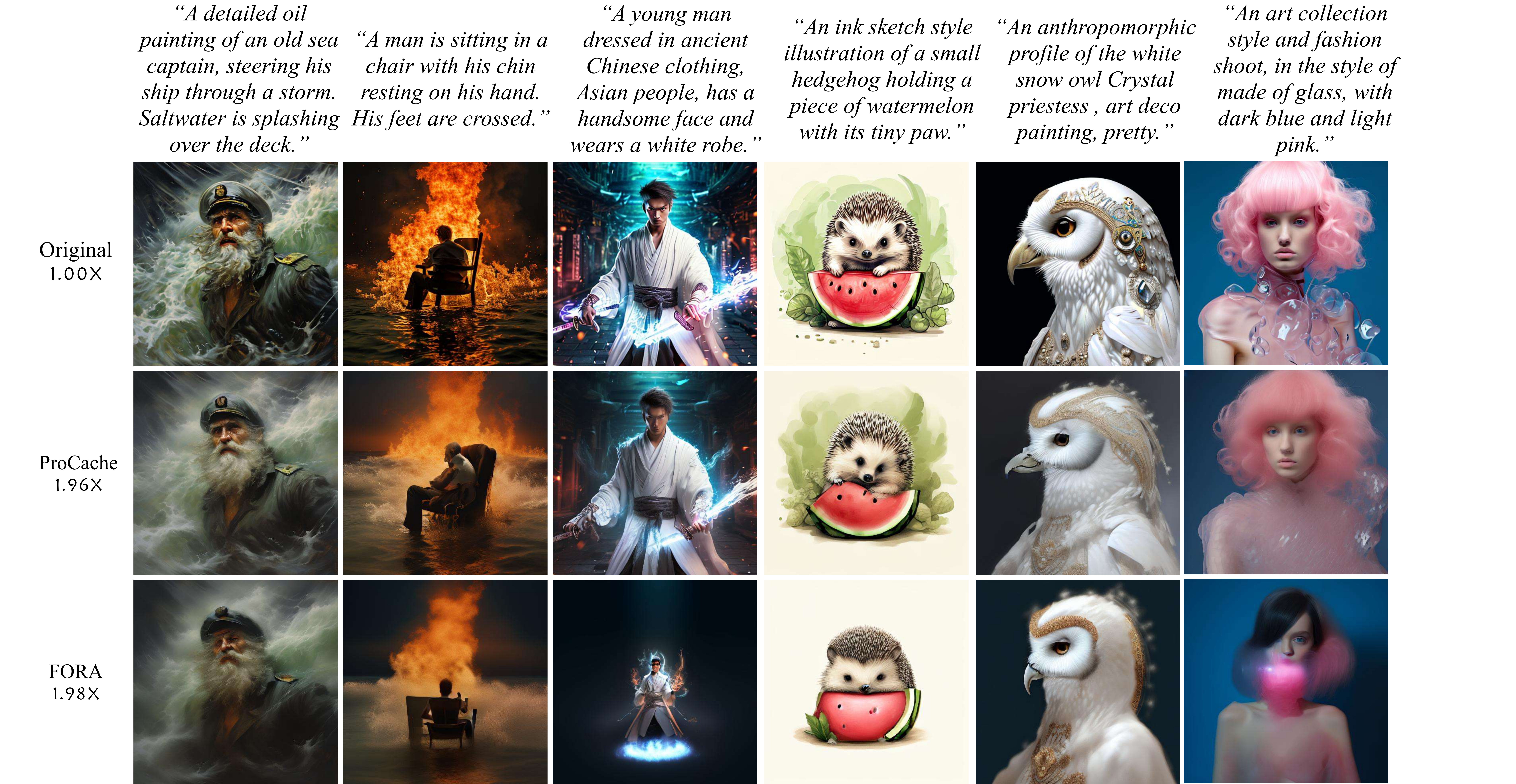}
\captionof{figure}{Visualization examples for different acceleration methods on Pixart-$\alpha$.}
\label{fig:vis_pix}
\end{minipage}
\end{figure}

To further illustrate the qualitative improvements of our method, we present visualization examples on Pixart-$\alpha$ and DiT (Figure \ref{fig:dit_apx}, Figure \ref{fig:vis_pix}, Figure \ref{fig:pix_apx}). With similar speed-up ratios compared with FORA \cite{fora} (1.96$\times$), our method produces more realistic, detailed images and better alignment with original images and text prompts. For example, under the prompt \textit{A young man dressed in ancient Chinese clothing, asian people, has a handsome face and wears a white robe.”}, the image generated by FORA exhibit significant visual degradation, affecting both the main subject and fine details. In contrast, ProCache preserves details effectively, retaining even subtle elements such as the blue light adjacent to the samurai’s body. These results showcase the superior fidelity and consistency of our method in generating high-quality outputs across diverse scenarios.

\begin{figure}[ht]
\centering
\includegraphics[width=0.85\textwidth]{figs/vis_pix_apx.pdf}
\caption{Visualization examples for different acceleration methods on Pixart-$\alpha$.}
\label{fig:pix_apx}
\end{figure}

\section{Visualization of Searched Caching Pattern}
\label{supp:Final Pattern Details}

The searched caching patterns for PixArt-$\alpha$, DiT-XL/2, and FLUX.1-dev are shown in Figure \ref{fig:final_patterns}. Unlike uniform caching strategies that apply fixed intervals throughout the denoising process, ProCache discovers non-uniform patterns that align with the temporal dynamics of diffusion transformers. Specifically, the patterns exhibit longer reuse intervals in early and middle denoising stages where feature changes remain minimal, and progressively shorter intervals in later stages where features evolve rapidly. This adaptive scheduling effectively balances computational efficiency with generation quality by allocating more computation resources to critical time periods while maximizing reuse during stable phases. ProCache achieves superior quality-speed trade-offs compared to uniform caching strategies.


\end{document}